\documentclass[12pt]{article}

\usepackage[superscript,biblabel]{cite}

\usepackage{graphicx}
\usepackage{tabularx}

\usepackage{times}

\title{Empirical observation of negligible fairness-accuracy trade-offs in machine learning for public policy}

\author
{Kit T. Rodolfa,$^{1}$ Hemank Lamba,$^{1}$ Rayid Ghani$^{1\ast}$\\  
\\
\normalsize{$^{1}$Machine Learning Department and Heinz College of Information}\\ 
\normalsize{Systems and Public Policy, Carnegie Mellon University,}\\
\normalsize{Pittsburgh, PA 15213, USA}\\
\\
\normalsize{$^\ast$To whom correspondence should be addressed; E-mail:  rayid@cmu.edu.}
}

\date{}

\begin{document} 




\maketitle 

\begin{abstract}

Growing use of machine learning in policy and social impact settings have raised concerns for fairness implications, especially for racial minorities. These concerns have generated considerable interest among machine learning and artificial intelligence researchers, who have developed new methods and established theoretical bounds for improving fairness, focusing on the source data, regularization and model training, or post-hoc adjustments to model scores. However, little work has studied the practical trade-offs between fairness and accuracy in real-world settings to understand how these bounds and methods translate into policy choices and impact on society. Our empirical study fills this gap by investigating the impact of mitigating disparities on accuracy, focusing on the common context of using machine learning to inform benefit allocation in resource-constrained programs across education, mental health, criminal justice, and housing safety. Here we describe applied work in which we find fairness-accuracy trade-offs to be negligible in practice. In each setting studied, explicitly focusing on achieving equity and using our proposed post-hoc disparity mitigation methods, fairness was substantially improved without sacrificing accuracy. This observation was robust across policy contexts studied, scale of resources available for intervention, time, and relative size of the protected groups. These empirical results challenge a commonly held assumption that reducing disparities either requires accepting an appreciable drop in accuracy or the development of novel, complex methods, making reducing disparities in these applications more practical.

\end{abstract}

There has been a rapid growth in the use of machine learning for applications with extensive impact on society, such as informing bail determination decisions \cite{Chouldechova2017FairInstruments,Skeem2016RiskImpact,Angwin2016MachineBias}, hiring \cite{Raghavan2020MitigatingPractices}, healthcare delivery \cite{Obermeyer2019DissectingPopulations,Ramachandran2020PredictiveClinic}, and social service interventions \cite{Bauman2018ReducingInterventions,Chouldechova2018ADecisions,Potash2015PredictivePoisoning}. These wide-reaching applications have been met with heightened concerns about their potential for introducing or amplifying inequities, especially for racial minorities and economically disadvantaged individuals, motivating exploration of a range of potential sources and mitigation strategies for biases, including in the underlying data \cite{Chen2018WhyDiscriminatory}, labels \cite{Obermeyer2019DissectingPopulations}, model training \cite{ElisaCelis2019ClassificationGuarantees,Zafar2017FairnessMistreatment,Dwork2018DecoupledLearning}, and post-modeling adjustments to scores \cite{Hardt2016EqualityLearning,Rodolfa2020CaseInterventions}. A common underpinning of much of this work is the assumption that trade-offs between equity and accuracy may necessitate complex methods or difficult policy choices \cite{Heidari2018FairnessMaking,Friedler2019ALearning,Kearns2019AnLearning,Zafar2017FairnessClassification}, however little work to date has explicitly evaluated the magnitude (or even the existence) of these trade-offs in real-world problems. Note that in this work we use the term ``accuracy'' in the more colloquial sense of the correctness of a model's predictions relative to the task at hand (in contrast to the fairness of those predictions), rather than the specific statistical property of the same name.  For each policy setting we study we use a more specific ``accuracy'' metric based on the goal of the program. 

Our study focuses on testing the assumed accuracy-fairness trade-offs in resource allocation problems across several public policy domains. Organizations with limited resources are often only able to intervene and allocate benefits to a relatively small number of individuals with need, presenting a ``top $k$'' optimization problem where model accuracy is judged by $precision$ (also known as positive predictive value) among the $k$ highest-scoring individuals (although precision in the top $k$ readily maps to a concept of efficiently allocating limited resources in these settings, it is worth noting that for a given value of $k$ on a fixed dataset, knowing the value of precision in the top $k$ will fully determine both the values of recall in the top $k$ and accuracy in the top $k$ such that optimizing for any one of these metrics is equivalent to optimizing for the other two as well). In such assistive intervention settings, we \cite{Rodolfa2020CaseInterventions} and others \cite{Hardt2016EqualityLearning} have argued that recall (also known as sensitivity or true positive rate) disparities are often an appropriate equity metric, reflecting a concept of ``equality of opportunity.'' In a recent case study \cite{Rodolfa2020CaseInterventions}, we found that \textbf {explicitly focusing on achieving equity and using subgroup-specific score thresholds as a post-hoc disparity mitigation method improved the equity of predictions with only a very modest decrease in accuracy}. While that study focused on our experiences incorporating fairness into the deployment of a machine learning system and the related policy decision-making in a single context (developing social service interventions as a means of jail diversion in Los Angeles, CA), the empirical work here extends that study's surprising result to several new policy contexts and modeling choices. Our results suggest that trade-offs between fairness and effectiveness can in fact be negligible in practice, suggesting that improvement in may be equity easier and more practical across a wide range of applications than is often expected. We come to this conclusion using a variety of projects we undertook over the past few years with government agencies across criminal justice, mental health, housing safety, and education, finding that each of these contexts poses a counterexample to the assumption of large trade-offs between fairness and accuracy.

\section*{Policy Settings and Data}

Information about the policy settings included in the present study are provided in Table \ref{tab:data-details}, and a brief description of the context and problem for each is provided below. Note that the machine learning formulation (including outcome measures, train-validation splits, evaluation metrics) and details in each case were determined through careful scoping and collaboration with the partner organization. For instance, the top $k$ list size in each case reflects the organization's resource constraints for intervening on entities, while policymaker and stakeholder priorities informed the label definition as well as the sensitive attribute used for measuring fairness.

\underline{Inmate Mental Health} Seeking to break the cycle of incarceration for individuals with untreated mental health conditions, Johnson County, KS, partnered with us to prioritize limited resources for mental health outreach on individuals at risk of a future jail booking. We developed a predictive model of risk for a booking in the next year, focusing on identifying 500 individuals for outreach in a 4-month window based on the resources available to the program. Disparities on race and ethnicity are particularly salient in the criminal justice context, and we focused on this attribute in our bias analyses (here, we look either at 2-way results between white and non-white individuals, or 3-way results between white, Black, and Hispanic individuals; individuals in other or unknown racial or ethnic groups make up only 0.4\% of the dataset and were included with white individuals for these analyses).

\underline{Housing Safety} The Code Enforcement Office in San Jose, CA, is tasked with protecting occupants of properties with multiple units (such as apartment buildings) by conducting safety inspections, but doesn't have sufficient staffing to inspect all 4,500 properties every year. Using internal data supplied by the program, we developed a model for the risk that a serious violation would be found if a given property were prioritized for inspection. We focused on disparities between housing units in higher- and lower-income neighborhoods (median income above or below \$55,000) where considerable disparities were observed in our initial models favoring higher-income areas.

\underline{Student Outcomes} El Salvador's Ministry of Education seeks to support students to reduce the country's substantial dropout rates (recently as high as 29\% in some years), but the budget for these programs is insufficient to reach every student. Student-level data was provided by the Ministry to develop a model of students at risk of dropping out, and our analysis here focuses on identifying the 10,000 highest-risk students in the state of San Salvador. The Ministry of Education was concerned with potential disparities in gender, age relative to grade level, and urban-rural divide. Initial analyses found large disparities with ``over-age'' students (those at least 2 standard deviations above the mean of their grade level), which we focus on in the present study.

Since the projects above used confidential and sensitive data and were done under data use agreements, we are not able to make that data publicly available. For our work to be easily reproducible, we include a fourth problem in this study where the data is available publicly: 

\underline{Education Crowdfunding} The non-profit DonorsChoose helps alleviate school funding shortages by providing a crowdfunding platform for teachers to post requests for their classroom needs. Here, we make use of a dataset DonorsChoose made publicly available in 2014 and posit an effort to assist projects at risk of going unfunded (for instance, providing a review and consultation) capable of helping 1,000 projects in a 2-month window. Reflecting the platform's goal of helping schools and teachers most at need, we focus in this context on disparities across school poverty levels (65\% free/reduced lunch vs others). Although unlike the other settings described above, the Education Crowdfunding analysis did not arise from a project undertaken in partnership with the organization, we sought to scope and formulate a project with the same characteristics of those we typically encounter with partners.

Details of the machine learning modeling and evaluation performed in each context are described in the Methods. In short, we explored an expansive grid of machine learning model types and hyperparameters in each problem to generate candidate models for our analysis. Because of the inherently non-stationary nature of real-world problems in general and policy applications in particular(with policies, practices, and context changing over time), training and validation sets were generated through a process of inter-temporal cross-validation. This approach creates a set of temporally sequential train and validation datasets, reflecting deployment of machine learning in these settings where models trained on past data must generalize into a changing future context. Although not the primary focus of the current work, we note that in each context, modeling was able to provide meaningful improvements over the base rate for the outcome (label) of interest in the population (shown in Table \ref{tab:data-details}), which could translate into substantive efficiency gains in policy implementation. Note that in a baseline model which randomly chose a group for intervention, the expected fraction of true positives in this group (that is, the precision) would be given by this base rate of the label. In each of the Housing Safety, Student Outcomes, and Education Crowdfunding problems, model performance provided a roughly 2-fold increase over this baseline, and in the Inmate Mental Health context (where the underlying event is more rare, with a base rate of 12\%), this improvement was by more than a factor of 4, making the ML models useful compared to simpler solutions and baselines. 

The following results focus on using group-specific score thresholds to mitigate disparities observed in these models' predictions. Fig. 1a illustrates the intuition behind making group-specific adjustments to a uniform score threshold in order to equalize recall (also known as sensitivity or true positive rate) across groups. Because recall increases monotonically with lower score thresholds, a unique solution can be found that equalizes recall across groups while keeping a fixed total number of entities selected for intervention. In Fig. 1b, we provide a schematic of the temporal strategy used for making these adjustments and testing their ability to generalize to unseen future data. In this figure, the dark-colored rectangles indicate points in time at which a cohort of training or validation examples is defined (this might be school years in the Student Outcomes context or quarters in the Housing Safety context) and the associated light-colored rectangles reflect a buffer during which outcomes are measured (for instance, the 4 months a project in the Education Crowdfunding context has to collect donations) in order to avoid leakage\cite{Top10Leakage} between the training and validation sets. Note that features for each cohort will make use of all earlier information as well. For one point in our analysis, a grid of models is trained using the blue cohort ($t_{-2}$), with performance on the purple cohort ($t_{-1}$) used to select well-performing models and find thresholds which equalize recall across groups. In order to evaluate how well these thresholds would then improve fairness (as well as any associated accuracy trade-off) when applied to new data, we step the process forward in time: training models using data up to the purple cohort and applying them along with the group-specific thresholds on the orange cohort ($t_{0}$), which reflects our true generalization performance. To understand consistency and stability of our results over time, we repeat this process several times, shifting the three cohorts together across different temporal validation splits.

\section*{Measuring Fairness}

One challenge encountered by both researchers and practitioners seeking to improve the fairness of machine learning models is the nebulous nature of the concept of ``fairness'' itself. Much has been written about the wide range of metrics which can be used to measure or conceptualize fairness in different contexts \cite{Verma2018FairnessExplained,Gajane2018Formalizing}, and, in particular, the mathematical incompatibility of various sets of these metrics \cite{Chouldechova2017FairInstruments,Kleinberg2017Inherent}. As a result, an important aspect of project scoping is understanding the fairness metric (or metrics) most appropriate to the given context. We have previously described a framework for this process \cite{Rodolfa2020CaseInterventions}. For instance, machine learning applications which inform benefit allocation decisions might be primarily concerned with avoiding disparities in false negatives (that is, failing to reach individuals with need), while applications that are more punitive in nature (such as bail denial decisions) are likely to be more concerned with disparities in false positives.

In the present work, we focus on resource-constrained assistive programs, a context in which we argue disparities in recall (also known as sensitivity or true positive rate) is a natural conceptualization of fairness. Because recall is the complement of the false negative rate (that is, $recall = TPR = 1-FNR$), improving equity in terms of recall reflects a focus on errors of omission, but provides more readily-interpretable ratios than the false negative rate itself when resources are limited (such that even a highly predictive model would have a false negative rate near 1). Hardt \cite{Hardt2016EqualityLearning} provides some additional helpful intuition here, describing this metric as a reflection of ``equality of opportunity.'' Specifically, when only a fraction of individuals with need can receive a benefit, recall measures the fraction of those individuals a program reaches and disparities in recall therefore measure whether individuals with need across different sub-groups have an equal chance of receiving the benefit. In particular, we measure recall disparity as the ratio between the recall attained by a model for one group of interest (such as Black individuals) and the model's recall for a reference group (such as white individuals). In contexts with more than two sub-groups of interest, we choose a reference group and consider disparities for all other groups relative to this group separately.

\section*{Evaluating Trade-Offs}

We started with the assumption, based on existing theoretical work, that there is a trade-off between ``accuracy'' and ``fairness.'' As an initial experiment, we explored how model ``accuracy'' changes upon adjusting for disparities in the Inmate Mental Health setting using a single temporal validation split (with validation set outcomes spanning 4/2018 to 4/2019). In Figs. 1c and 1d, each pair of points is a model specification: with results obtained without adjusting for equity in blue and those with the equity adjustment in orange. The x-axis shows the precision (positive predictive value) of each model on the 500 selected individuals and the y-axis shows the recall disparity between white and non-white individuals. Fig. 1c shows all models considered (the model specifications used here are detailed in Supplementary Table 1), while Fig. 1d provides a more detailed view of the better-performing models that might reasonably be selected. All of the unadjusted models had significant disparities, indicating that merely measuring disparities to inform the process of model selection would be insufficient for achieving fairness. However, applying our proposed bias mitigation method to adjust for disparities, we find little evidence of a fairness-accuracy trade-off: overall, the mean change in precision after adjustment is -0.0006 (std: 0.0087); Fig. 1e shows the distribution of these shifts.


We also investigated creating a composite model (based on the work of Dwork and colleagues \cite{Dwork2018DecoupledLearning}) by choosing the best-performing model for each subgroup and using the recall-equalizing set of individuals (maintaining a total list size of 500) from each subgroup-optimized model. The performance of this composite on the subsequent validation set is shown as a red diamond in Figs. 1c and 1d. On both fairness and accuracy-related metrics, this composite appears to perform competitively with other models, but does not stand out as out-performing the fairness-adjusted individual models on either metric. The promise of the composite strategy is that it might improve the accuracy on each subgroup (and, hence, the overall performance across subgroups) while yielding more equitable results. In this initial experiment, however, the more complex approach of building a composite model shows no indication of providing gains beyond the simpler approach of making fairness-enhancing adjustments to a single, well-performing model.

\section*{Comparing Mitigation Strategies}

These initial results suggested disparity mitigation could be integrated into the process of model selection, and we next sought to compare strategies for doing so (summarized in Table \ref{tab:seln-strategies} and described in more detail in the Methods). In the strategy labeled  ``Mitigated - Single Model,'' model specifications are compared based on their precision at top $k$ after applying group-specific thresholds to mitigate disparities and the chosen model is evaluated on new data. Selecting a model without regard to fairness and applying disparity mitigation only to the chosen model showed no practical difference in performance on either fairness or accuracy metrics in any of our analyses (see  Supplementary Discussion). As a baseline, the ``Unmitigated'' approach performs model selection without accounting for any disparities. 

Additionally, the ``Mitigated - Composite Model'' strategy chooses the best-performing model on each subgroup at the model selection stage and picks recall-balancing thresholds across these. We also explored composite models created from fully-decoupled models trained only on subgroup-specific examples (also suggested by Dwork \cite{Dwork2018DecoupledLearning}), but saw no difference in performance from the composite method described here in these initial experiments and did not pursue this line of investigation further. Likewise, in considering other fairness-enhancing methods that have been proposed, we found that some (such as the regularization method developed by Zafar \cite{Zafar2017FairnessMistreatment}) were not well-suited to the ``top k'' problem setting and others (such as the methods proposed by Celis \cite{ElisaCelis2019ClassificationGuarantees} and Menon \cite{KrishnaMenon2018TheClassification}) could be shown to be equivalent to group-specific thresholds under certain conditions (see Supplementary Discussion).    


Applying our three strategies to a wide variety of models and hyperparameter combinations including random forests, logistic regression, boosting, and decision trees built for each policy problem, \textbf{appreciable recall disparities were present in the unadjusted models in all cases}, ranging from 50\% higher recall favoring white individuals in the Inmate Mental Health context to as much as a 250\% disparity in the Student Outcomes setting. In the results shown in Fig. 2, strategies yielding larger disparities will have higher values along the y-axis while strategies yielding precision decreases would move left along the x-axis relative to the unmitigated models (solid squares marked with a 'U'). Note that in each context, the top $k$ list size and sensitive attribute for measuring performance and disparities, respectively, can be found in Table \ref{tab:data-details}, reflecting the resource constraints and stakeholder priorities of the policy settings discussed above. However, we see that \textbf{mitigating disparities, surprisingly, does not come with any appreciable degradation of overall model performance} on unseen data for any of the policy settings investigated: precision at top $k$ is similar in magnitude for the unmitigated and mitigated models, with little difference between the fairness-enhancing approaches (large error bars in the unmitigated Student Outcomes result reflect the small number of temporal splits and a single cohort with low baseline disparity). Across all four problems, the precision at top $k$ for the mitigated models was statistically indistinguishable from that of the unmitigated models (t-tests for these differences yielded p-values of 0.83 for Education Crowdfunding, 0.47 for Housing Safety, 0.84 for Inmate Mental Health, and 0.71 for Student Outcomes), as well as practically negligible with no difference in precision higher than 1 percentage point. This finding empirically demonstrates that \textbf{any inherent trade-off that might be present in adjusting for disparities appear to be practically non-existent}.


To explore these results in more detail, Fig. 3 shows the effect of these bias mitigation strategies on the Inmate Mental Health setting over temporal validation cohorts (see Supplementary Figs. 2-4 for results from the other policy settings). In this context, we also wanted to understand how these methods performed when adjusting for disparities across multiple subgroups, looking at race/ethnicity across white, Black, and Hispanic individuals.

Overall performance (precision at top 500) was similar for models selected from all three strategies over time (Fig. 3a) and recall disparities between white and Black individuals were consistently and significantly improved by the adjustments (Fig. 3b). Without accounting for equity or fairness, recall for white individuals was 40-100\% higher than Black individuals in the chosen models, while both fairness enhancing strategies yielded a ratio near one (parity) for these two subgroups. The results for disparities between Hispanic and white individuals (Fig. 3c) are broadly consistent, but show significantly more variation, which appears to be particularly acute with the composite model approach. We hypothesized that this variability might arise in part from an interaction between the relatively small size of this group in the population (about 11\% of each cohort) and process by which the composite model is created: while the model selection process for the single model approach makes use of the full top $k$ set of individuals selected for intervention, the composite approach performs model selection on each group separately, which may be less robust as the groups become smaller.

To better understand the interaction between the overall top $k$ list size, subgroup sizes, and our results, we performed a series of sensitivity experiments. Although in practice, each policy problem comes with a specific value of $k$ determined by the resource constraints of the organization taking action. Fig. 4a-c shows the fairness and accuracy metrics for each strategy at different levels of program resource availability ($k$). Consistent with the findings above, Fig. 4c shows essentially identical precision at top $k$ performance across all selection strategies, both bias-mitigated and unmitigated, and over all values of $k$ explored. Across all $k$, the fairness improving strategy choosing a single model consistently showed improvement in disparity metrics for both Black (Fig. 4a) and Hispanic (Fig. 4b) individuals. By contrast, the composite strategy was less effective at equalizing recall across race/ethnicity subgroups, particularly at lower list sizes and with the smaller Hispanic subgroup. 

The under-performance of the composite model suggests a mechanism driving this result: evaluating a model's performance on such a small group may be especially prone to overfitting, choosing a model with an unreasonably high estimate of precision on the subgroup that won't generalize well into the future. Because constructing the composite model across subgroups involves determining the number of individuals to select from each subgroup with the goal of equalizing recall across them, a process that leads to systematically over-estimating the precision for Hispanic individuals would bias towards selecting a smaller set of individuals than is actually needed (this can be observed in Supplementary Fig. 5b, which shows the fraction of Hispanic individuals in the selected list). When the performance of the chosen model fails to generalize well on the unseen, future data, the under-estimated subgroup size results in a lower-than-expected recall and, thus, a higher disparity. By contrast, choosing a single model across all groups (rather than building a composite) seems likely to be more resilient to this issue, both by virtue of reducing variance in the model selection process (by way of the larger overall sample sizes) and to the degree that any potential reduction in generalization performance might be likely to affect all subgroups similarly.

To further investigate the impact of subgroup size on the stability of the results (e.g., through sampling variation), we performed a resampling experiment to progressively increase the Hispanic fraction in the population, ranging from 5\% to 38\% (focusing on $k=500$ based on the actual program resources). Note that although this experiment focused on changing the fraction of Hispanic individuals in the population, the analysis made use of data from all subgroups, keeping the ratio between Black and white individuals constant (see Supplementary Fig. 6 for additional results). As observed above, the composite model performs less well at reducing disparities between white and Hispanic individuals for the baseline Hispanic fraction of 11\% (Fig. 4d). While this under-performance is also observed at other low fractions, this strategy becomes competitive with the other the single model strategy at Hispanic fractions above 20\%, consistent with our hypothesis that the small population of Hispanic individuals in the underlying data may be creating a tendency towards overfitting in the formation of the composite models.

\section*{Discussion}

Taken together, these results suggest a promising and novel conclusion: across a range of policy domains, mitigating disparities does not inherently require new and complex machine learning methods or a prohibitively large accuracy sacrifice as is often assumed. Instead, explicitly defining the fairness goal upfront in the machine learning process and making design choices to achieve that goal by taking active, practical steps, such as the post-hoc bias mitigation strategies investigated here, are important steps to achieving that goal. While it is of course important to keep in mind the empirical nature of this finding, we see an important contribution of this work in providing practical counterexamples to commonly-held assumptions about the nature of the relationship between fairness and accuracy. As such, we see the consistency of these results across policy settings and contexts here as strongly suggestive that practitioners applying machine learning methods to inform other high-stakes policy decisions should question this assumption in their own work. Additionally, this work contributes to a growing body of evidence that, in practice, straightforward approaches such as thoughtful label choice \cite{Obermeyer2019DissectingPopulations}, model design \cite{Chouldechova2018ADecisions}, or post-modeling mitigation can effectively reduce biases in many machine learning systems. 

A detailed understanding of the mechanisms driving the nominal trade-offs observed here is beyond the scope of the current empirical study, but two factors that we posit may play important roles are the resource-constrained top $k$ setting and the relative predictive performance of the models across subgroups. With regards to the top $k$ context, the relatively small number of individuals who can be selected for intervention will translate into a relatively high score threshold where examples with positive labels are relatively dense both just above and just below the threshold. Because reducing recall disparities involves trading true positives from one subgroup for true positives from another, operating at a region of the score distribution where positive labels are relatively dense means that the perturbations to group-specific score thresholds required to eliminate a given level of disparity may be reasonably small. Here, our hypothesis is that settings in which resources are less constrained may pose a greater challenge to disparity mitigation, because operating deeper in the score distribution (that is, at lower predicted score thresholds) means the recall gradient will typically be flatter and larger adjustments may be required to achieve equity (and, hence, resulting in greater trade-offs). Likewise, the relative performance of the model across subgroups will be a key factor in determining both how much the score thresholds will need to change for each group to collect enough true positives to achieve recall equity as well as how many false positives are included in the process of doing so. Contexts in which the available features are much less predictive of the outcome for some subgroups relative to others could lead to both larger baseline disparities (if recall at a given threshold is higher for the more readily-predicted group) and more significant fairness-accuracy trade-offs (by necessitating larger adjustments to achieve recall equity). Although these two factors may be more salient in other settings, our focus on  high-impact benefit allocation programs with limited resources reflects a setting that occurs very commonly in policy contexts, and frequently encountered in our applied work \cite{Potash2015PredictivePoisoning,Rodolfa2020CaseInterventions,Ramachandran2020PredictiveClinic,Bauman2018ReducingInterventions} and the empirical results here suggest some of these concerns may not be prohibitively common in practice. Nevertheless, an important avenue for future work will be further exploring these hypothesis and beginning to develop a better theoretical understanding of the conditions under which improvements in equity will or will not come at appreciable expense in accuracy.

Much has also been written about the wide variety of fairness metrics that may be relevant depending on the context \cite{Chouldechova2017FairInstruments,Hardt2016EqualityLearning,Rodolfa2020CaseInterventions,Verma2018FairnessExplained,Gajane2018Formalizing}, and further exploration of the fairness-accuracy trade-offs in those contexts is certainly warranted, particularly where balancing multiple fairness metrics may be desirable. Likewise, it may be possible that there is a tension between improving fairness across different attributes (e.g., sex and race) or at the intersection of attributes. Future work should also extend these results to explore the impact not only on equity in decision making, but also equity in longer-term outcomes and implications in a legal context such as those discussed by Huq \cite{Huq2019RacialJustice}.

\subsection*{Social Impact of This Work}
Beyond the immediate implications for the policy and social impact problems explored here, our goal in presenting these empirical findings here is to better inform the use of machine learning in high-stakes decisions by challenging the commonly held belief that there necessarily is a trade-off between ``accuracy'' and ``fairness'' in these applications. However, it is also important to note that fairness is not only a function of a model's predictions but also how those predictions are acted on by human decision-makers and, more broadly, the context in which it operates, with historical, cultural, and structural sources of inequities that society as a whole must strive to overcome through the ongoing process of remaking itself to better reflect its highest ideals of justice and equity. Improving the fairness of the machine learning models that continue to find growing applications in critical decisions that affect many aspects of people's lives may be only one small element of that process, but we hope this work will inspire researchers, policymakers, and data science practitioners alike to explicitly consider fairness as a goal and take steps, such as those proposed here, in their work that can, collectively, contribute to bending the long arc of history towards a more just and equitable society.

\section*{Methods}

\noindent \textbf{Policy Contexts and Data}

A key aim of this work was to explore the fairness-accuracy trade-offs encountered in practice in the context of machine learning applications for public policy settings. As such, we drew on several projects we have worked on in partnership with government agencies across policy domains. We describe these contexts briefly in the main text and provide more details about each setting below:

\underline{Inmate Mental Health} Untreated mental health conditions often result in a negative spiral, which can culminate in repeated periods of incarceration with long term consequences both for the affected individual and the community as a whole \cite{Hamilton2010Needs}. Surveys of inmate populations have suggested a high prevalence of multiple and complex needs, with 64\% of people in local jails suffering from mental health issues and 55\% meeting criteria for substance abuse or dependence \cite{James2006Inmates}. The criminal justice system is poorly suited to address these needs, yet houses three times as many individuals with serious mental illness as hospitals \cite{FullerTorrey2010Survey}. In 2016, Johnson County, KS, partnered with our group to help them break this cycle of incarceration by identifying individuals who might benefit from outreach with mental health resources and are at risk for future incarceration. While the Johnson County Mental Health Center (JCMHC) currently provides services to the jail population, needs are generally identified reactively, for instance through screening instruments individuals fill out when entering jail. The new program being developed will supplement these existing approaches by adding a new automatic referral system for people who are at risk of being booked into jail, with the hope that they can be outreached before they return to jail. Through our partnership, the county provided administrative data from their mental health center, jail system, police arrests, and ambulance runs. Modeling was focused on a cohort of Johnson County residents with any history of mental health need who had been released from jail within the past three years. Early results from this work were described previously\cite{Bauman2018ReducingInterventions}. A field evaluation of the predictive model is ongoing at the time of this writing, but validation on historical data demonstrated a 12\% improvement over a baseline based on the number of bookings in the prior year and 4.8-fold increase over the population prevalence.

\underline{Housing Safety} The Multiple Housing team in San Jose's Code Enforcement Office is tasked with protecting the occupants of properties with three or more units, such as apartment buildings, fraternities, sororities, and hotels. They do so by conducting routine inspections of these properties, looking for everything from blight and pest infestations to faulty construction and fire hazards (see work by Holtzen\cite{Holtzen2016Housing} and Klein \cite{Klein2015Housing} for a discussion of the importance of housing inspections to public health). Although the city of San Jose inspects all of the properties on its Multiple Housing roster over time, and expects to find minor violations at many of them, it is important that they can identify and mitigate dangerous situations early to prevent accidents. With more than 4,500 multiple housing properties in San Jose, CA -- many of which comprise multiple buildings and hundreds of units -- it is not possible for the city to inspect every unit every year. San Jose recently instituted a tiered approach to prioritizing inspections, inspecting riskier properties more frequently and thoroughly. Although the tier system helped focus inspections on riskier properties, the new system has its limitations. The city evaluates tier assignments for properties infrequently (every 3 to 6 years), and these adjustments require a great deal of expertise and manual work while leaving out a rich amount of information. In order to provide a more nuanced view of properties' violation risk over time and allow for more efficient scheduling of inspections, the Code Enforcement Office partnered with us to develop a model to predict the risk that a serious violation would be found if a given property was prioritized for inspection (similar tools have been developed for allocating fire inspections in New York \cite{Athey2017BeyondPrediction} and health inspections in Boston \cite{Glaeser2016Housing}). Evaluation of the model on historical data indicated that it could provide a 30\% increase in precision relative to the current tier system and the model's predictive accuracy was confirmed during a 4-month field trial in 2017.

\underline{Student Outcomes} Each year from 2010 through 2016, 15-29\% of students enrolled in school in El Salvador did not return to school in the following year. This high dropout rate is cause for serious concern, with significant consequences for economic productivity, workforce skill, inclusiveness of growth, social cohesion, and increasing youth risks \cite{Levin2007Price,Atwell2019GradNation}. El Salvador's Ministry of Education has programs available to support students with the goal of reducing these high dropout rates, but the budget for these programs is not large enough to reach every student and school in El Salvador. Predictive modeling has been deployed to help schools identify students at risk of dropping out in several contexts \cite{Lakkaraju2015Students,Aguiar2015Who,Bowers2012DropOut} and El Salvador partnered with us in 2018 to make use of these methods to focus their limited resources on the students at highest risk of not returning each year. Student-level data was provided by the Ministry of Education, including demographics, urbanicity, school-level resources (e.g., classrooms, computers, etc), gang and drug violence, family characteristics, attendance records, and grade repetition. For the present study, we focused on the state of San Salvador and identifying the 10,000 highest-risk students, considering annual cohorts of approximately 300,000 students and drawing on 5 years' of prior examples as training data.

\underline{Education Crowdfunding} Many schools in the United States, particularly in poorer communities, face funding shortages \cite{Morgan2018Gaps}. Often, teachers themselves are left to fill this gap, purchasing supplies for their classrooms when they have the individual resources to do so \cite{Hurza2015Teachers}. The non-profit DonorsChoose was founded in 2000 to help alleviate these shortages by providing a platform where teachers post project requests focused on their classroom needs and community members can make individual contributions to support these projects. Since 2000, they have facilitated \$970 million in donations to 40 million students in the United States \cite{DonorsChooseAbout}. However, approximately one third of all projects posted on the platform fail to reach their funding goal. Here, we make use of a dataset DonorsChoose made publicly available for the 2014 KDD Cup (an annual data mining competition) including information about projects, the schools posting them, and donations they received. Because the other case studies explored here focused on proprietary and often sensitive data shared with us under data use agreements that cannot be made publicly available, we included a case study surrounding this publicly-available dataset. While we have not partnered with DonorsChoose to deploy the machine learning system described, we otherwise treated this case study as we would any of our applied projects. Here, we consider a resource-constrained effort to assist projects at risk of going unfunded (for instance, providing a review and consultation) capable of helping 1,000 projects in a 2-month window, focusing on the most recent 2 years' of data available in the extract (earlier data had far fewer projects and instability in the baseline funding rates as the platform ramped up). This dataset is publicly available at kaggle.com \cite{DonorsChooseData}. \\

\noindent \textbf{Machine Learning Details}

All machine learning models, including feature engineering, model training, and performance evaluation were run using our open-source \verb|python| ML pipeline package, \verb|triage|. Machine learning methods used are from \verb|sklearn| (a python package) or \verb|catwalk| (a component of \verb|triage| for baselines methods as well as \verb|ScaledLogisticRegression|, which wraps the \verb|sklearn| logistic regression to ensure input features/predictors as scaled between 0 and 1). The modeling grid for each project is described in Supplementary Tables 1-4, reflecting the modeling space explored by the teams working on each project. For each estimator in the tables, the grid search considered reflects the full cross-product of the hyperparameter values specified. Here we make use of a variety of state-of-the-art machine learning methods for binary classification problems. Although precision in the top $k$ is the metric of interest in all of the settings discussed here (as described in the main text), few methods have been developed that seek to optimize for this metric directly. Instead, we are concerned with the relationship between fairness (measured here in terms of recall disparities) and accuracy (measured here in terms of precision in the top $k$) through real-world settings in practice. As such, we make use of well-established methods that are widely applied in practical settings, which themselves optimize for a variety of underlying target metrics (such as minimizing the regularized logistic loss in logistic regression or maximizing the information gain at each split in tree-based models). By training a large grid of estimator types and hyperparameter values, optimization for performance in terms of precision in the top $k$ can then be performed through the process of model selection over validation sets. As illustrated in Fig. 1b, we used a strategy of inter-temporal cross-validation (as described by Roberts \cite{Roberts2017Temporal} and Ye \cite{Ye2019Tenants}) to ensure that model evaluation and selection was done in a manner that reflected performance on novel data while guarding against ``leakage'' of information from the future affecting past results.

The method we used for mitigating disparities by post-modeling adjustment involving choosing sub-group specific thresholds (see Fig. 1a) was described in detail in our previous case study\cite{Rodolfa2020CaseInterventions} and draws on the idea of ``equality of opportunity'' discussed by Hardt\cite{Hardt2016EqualityLearning}. In brief, because the notion of fairness relevant in these policy settings relies on equalizing recall across groups, and recall monotonically increases with depth traversed in a model score, unique score thresholds that balance recall across groups can be readily found for a given combined list size. For each model, we calculate within-group recall values up to each individual in an initial validation set (purple / $t_{-1}$ in Fig. 1b), order the combined set by within-group recall and take the top $k$ individuals from this reordered set, calculating $k_{g}$ for each group $g$ such that $\sum{k_{g}} = k$ (the total top $k$ list size desired) and recall is balanced across groups. To evaluate this process on novel data, models were tested on a future cohort (tan / $t_{0}$ in Fig. 1b) and the top $k_{g}$ examples (ranked by score, then randomly to break ties) from each sub-group were selected to measure precision at top $k$ and recall disparities. In the process of model selection, we explored applying these disparity-mitigating thresholds either before choosing a model specification (``Mitigated - Single Model'') or after (``Mitigated - Unadj. Model Seln.''), finding no substantive difference in performance (Supplementary Fig. 1). For the ``Mitigated - Composite Model'' approach, a similar method was used, but within-group precision up to each individual is calculated for each model as well to determine the best model specification for each sub-group at each list depth (drawing on the ideas suggested by Dwork and colleagues\cite{Dwork2018DecoupledLearning}), then $k_{g}$ and group-specific model specifications are chosen for evaluation on novel data.

Code for all four projects including \verb|triage| configuration files specifying the full feature sets used as well as code used to mitigate disparities and evaluate fairness is available at \verb|github.com/dssg/peeps-chili|\\

\noindent \textbf{Data Availability} 

Data from the Inmate Mental Health context was shared through a partnership and data use agreement with the county government of Johnson County, KS (which collected and made available data from the county- and city-level agencies in their jurisdiction as described in the Methods above). Data from the Housing Safety context was shared through a partnership and data use agreement with the Code Enforcement Division in the city of San Jose, CA. Data from the Student Outcomes setting was shared through a partnership and data use agreement with the Ministry of Education in El Salvador. Although the sensitive nature of the data for these three contexts required that the work was performed under strict DUAs and the data cannot be made publicly available, researchers or practitioners interested in collaborating on these projects or with the agencies involved should contact the corresponding author (rayid@cmu.edu) for more information and introductions. The Education Crowdfunding dataset, however, is publicly available at: kaggle.com/c/kdd-cup-2014-predicting-excitement-at-donors-choose. Additionally, a database extract with model outputs and disparity mitigation results using this dataset is available for download (see replication instructions in the code repository noted below).\\

\noindent \textbf{Code Availability} 

The code used here for modeling, disparity mitigation, and analysis for all four projects is available at github.com/dssg/peeps-chili.\cite{PeepsChili} Complete instructions for replication of the Education Crowdfunding results reported here can be found in the README of this respository, along with a step-by-step jupyter notebook to perform the analysis.\\

\section*{Acknowledgments}
We would like to thank the Data Science for Social Good Fellowship fellows, project partners, and funders as well as our colleagues at the Center for Data Science and Public Policy at University of Chicago for the initial work on projects that were extended and used in this study. We also thank Kasun Amarasinghe for helpful discussions on the study and drafts of this paper. Parts of this work were funded by the National Science Foundation under grant IIS-2040929 (KTR, RG) and by a grant (unnumbered) from the C3.ai Digital Transformation Institute (KTR, HL, RG).

\section*{Author Contributions Statement} 
KTR: Conceptualization, Methodology, Software, Investigation, Writing - Original Draft; HL: Investigation, Software, Writing - Review \& Editing; RG: Conceptualization, Supervision, Funding acquisition, Writing - Review \& Editing 

\section*{Competing Interests Statement} 
The authors have no conflicts of interest to declare.


\clearpage

\begin{table}[htb!]
 \caption{Policy Settings and Data Details}
  \centering
  \newcolumntype{Y}{>{\centering\arraybackslash}X}
  \begin{tabularx}{1.1\linewidth}{|l|Y|Y|Y|Y|}
    \hline
     & \textbf{Inmate Mental Health} & \textbf{Housing Safety} & \textbf{Student Outcomes} & \textbf{Education Crowdfunding}  \\
    \hline
    Prediction Task & Jail booking within the next 12 months & Housing unit having a violation within the next year & Student not returning to school next year & Project not getting fully funded within 4 months \\ 
    \hline
    Timespan & 2013-01-01 to 2019-04-01 & 2011-01-01 to 2017-06-01 & 2009-01-01 to 2018-01-01 &  2010-01-01 to 2014-01-01 \\ 
    \hline
    \# of Entities & 61,192 & 4,593 & 801,242 & 210,310 \\
    \hline
    \# of Features & 3,465 & 1,657 & 220 & 319  \\ 
    \hline
    Base Rate & 0.12 & 0.43 & 0.25 & 0.24  \\ 
    \hline
    Evaluation Metric & Precision at top 500 & Precision at top 500 & Precision at top 10,000 & Precision at top 1,000  \\ 
    \hline
    Sensitive Attribute & Race & Median Income & Age Relative to Grade & Poverty Level  \\
    \hline
  \end{tabularx}
  \label{tab:data-details}
\end{table}

\clearpage

\begin{table}[htb!]
 \caption{Descriptions of Model Selection Strategies}
  \centering
  \def\arraystretch{1.5} 
  \begin{tabularx}{\linewidth}{ l X }
    \hline
    Strategy     & Description  \\
    \hline
    Unmitigated & Baseline strategy with no equity adjustments \\
    Mitigated - Composite Model & The highest-precision model is chosen for each subgroup and a composite model is formed combining these, with equity-balancing thresholds used when calculating test performance \\
    Mitigated - Single Model & All models are adjusted for recall equity and then the best model is selected based on precision and equity-balancing thresholds are applied for calculating test performance \\
    Mitigated - Unadj. Model Seln. & Model is selected based on precision then equity-balancing thresholds are applied for calculating test performance (See the Supplementary Discussion for results from this approach) \\
    \hline
  \end{tabularx}
 \label{tab:seln-strategies}
\end{table}

\clearpage


\begin{figure*}[htb!]
  \centering
  \includegraphics[width=\linewidth]{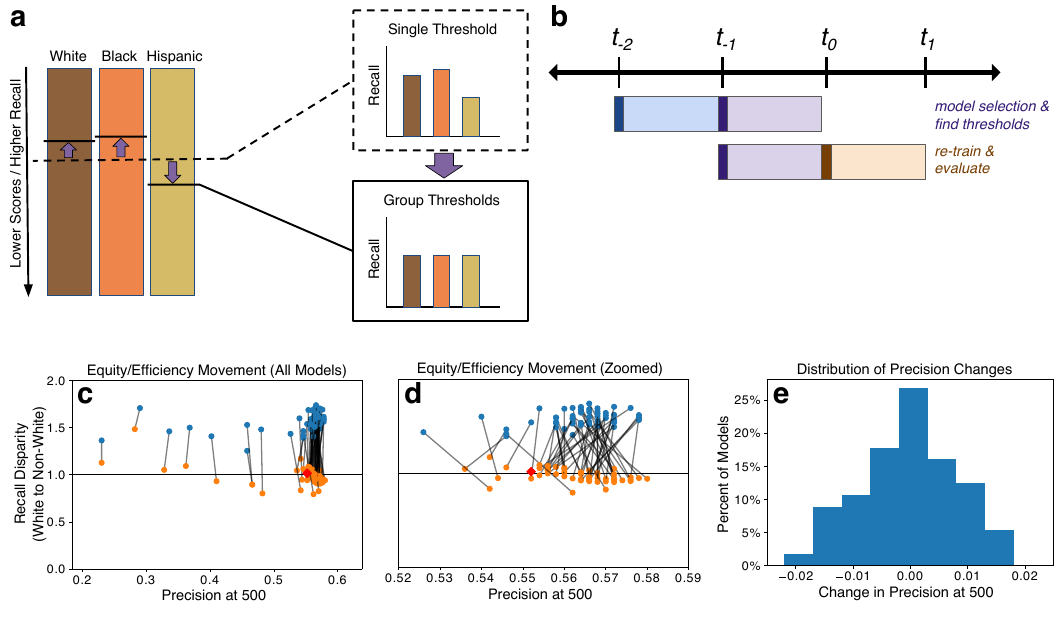}
  \caption{Illustration of the methods used and motivating results. (a) Subgroup-specific thresholds are applied to a modeled risk score to improve the recall equity among individuals chosen for intervention while maintaining a desired overall list size. (b) Temporal validation strategy: a grid of models is trained using examples as of $t_{-2}$ (dark blue, with labels derived from the time shown in light blue) and predictions on a cohort as of $t_{-1}$ (dark purple with labels derived from the time shown in light purple) are used to determine the equity-balancing thresholds described in (a). Models are then re-trained on this cohort for ``current day'' predictions as of $t_{0}$ (dark tan, with labels in light tan) used to evaluate model performance with equity adjustments. (c and d): Changes in race/ethnicity recall disparities before (blue) and after (orange) making post-hoc score adjustments for fairness in the Inmate Mental Health context. (c) shows all model specifications and (d) shows the cluster of well-performing models. The red diamond reflects the performance of a composite model combining the best-performing model for each subgroup. (e) Distribution of precision changes after adjusting for disparities for the models shown in (c), relative to the precision attained by the same model specification without adjustment (that is, the difference along the x-axis of the blue and orange dots).}
  \label{fig:methods}
\end{figure*}



\clearpage


\begin{figure*}[htb!]
  \centering
  \includegraphics[width=\linewidth]{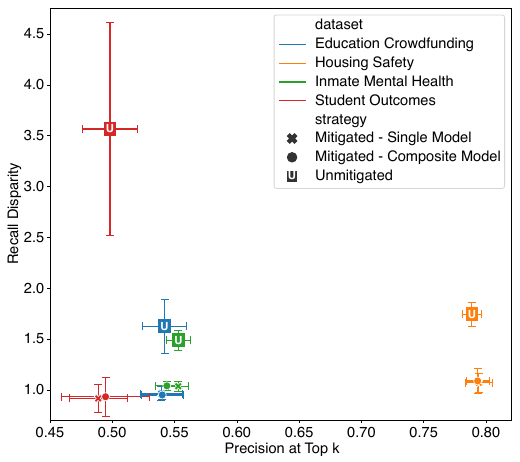}
  \caption{Comparing equity (recall disparity) and performance (precision at top $k$) metrics for different model selection strategies between different policy contexts. In the Education Crowdfunding context, models are evaluated at $k=1000$ across 10 temporal validation cohorts; in the Inmate Mental Health context, $k=500$ across 10 temporal validation cohorts; in the Student Outcomes context, $k=10,000$ across 5 temporal validation cohorts; and in the Housing Safety context, $k=500$ across 9 temporal validation cohorts. Unmitigated (baseline) models are shown as solid squares marked with a `U'. Decreases of strategies involving disparity mitigations relative to the y-axis demonstrate improvements in equity while showing little or no decrease in overall performance (that is, leftward movement on the x-axis). Error bars reflect 95\% confidence intervals across temporal validation cohorts.}
  \label{fig:compare}
\end{figure*}

\clearpage


\begin{figure*}[htb!]
  \centering
  \includegraphics[width=0.8\linewidth]{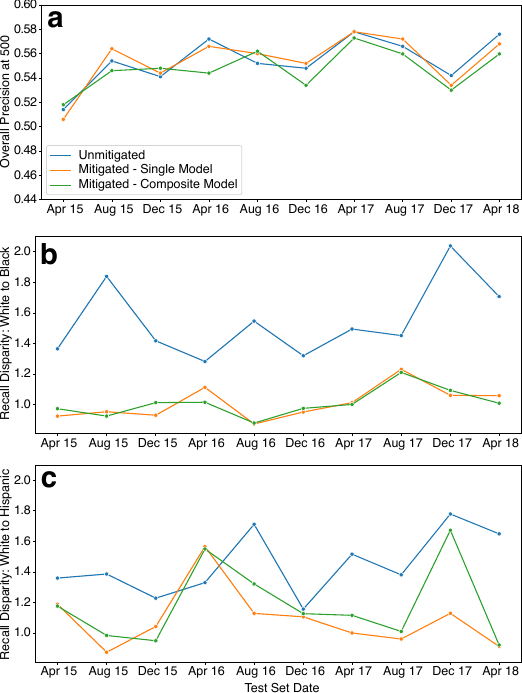}
  \caption{Comparing disparity and performance metrics over time for different model selection strategies. Results from the Inmate Mental Health policy setting using a total list size $k=500$. (a) Model performance in terms of overall precision at top 500. (b) Recall disparities between white and Black individuals. (c) Recall disparities between white and Hispanic individuals.}
  \label{fig:seln-strategy}
\end{figure*}

\clearpage


\begin{figure*}[htb!]
  \centering
  \includegraphics[width=\linewidth]{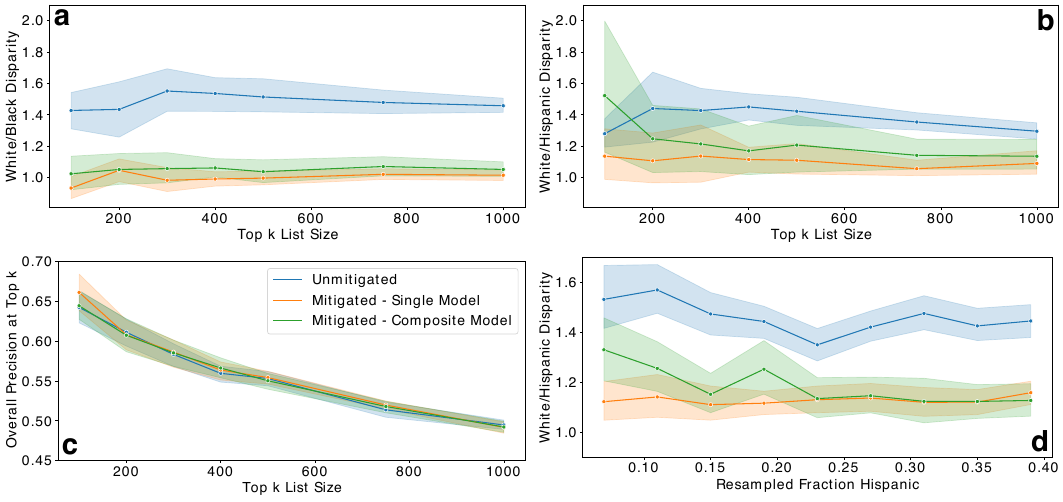}
  \caption{Comparing disparity and performance metrics across program scale and protected group size in the Inmate Mental Health policy setting. (a to c) Variation in results by list size: (a) white/Black disparities, (b) white/Hispanic disparities, and (c) overall precision at top $k$. (d) Disparities between white and Hispanic individuals by resampled size of the protected group in the overall population (using a list size of $k=500$). Shaded intervals reflect 95\% confidence intervals from variation across temporal validation splits (a-c) as well as bootstrap samples in (d).}
  \label{fig:list-size}
\end{figure*}

\end{document}





\maketitle 







\section*{Supplementary Discussion}
The results discussed in the main text focus on post-hoc score adjustment methods, drawing on several previous lines of work\cite{Dwork2018DecoupledLearning,Hardt2016EqualityLearning,Rodolfa2020CaseInterventions} and finds these straightforward methods suffice to significantly reduce disparities with little accuracy trade-off in the settings considered. However, we also did some preliminary work exploring a number of other methods that have been proposed, particularly to enhance fairness during the modeling process itself. In public policy settings with limited resources, regularization methods (such as that described by Zafar and colleagues\cite{Zafar2017FairnessMistreatment}) provide a particular challenge. These techniques typically seek to find the best overall classifier subject to some fairness constraint. This may be well-suited to contexts where there are no hard constraints on the resources available to act on predictive positives (for instance, in deciding whether to order a relatively low-cost medical diagnostic test). 

However, many applications, particularly in public policy contexts, are subject to the further constraint of limited resources --- these applications are best formulated as ``top-k'' problems, but unfortunately this formulation introduces a non-convex optimization that is not readily integrated into current fairness-constrained methods. Likewise, naively thresholding the resulting score from these methods to yield a set number of predicted positives provides no guarantee that the fairness constraints will hold. Supplementary Fig. 7 provides an example with data from the Inmate Mental Health context: using the method described by Zafar\cite{Zafar2017FairnessMistreatment} to generate a score with a false negative rate fairness constraint (note that $FNR = 1-TPR$, so this constraint is equivalent to recall equity) and choosing the top 500 individuals performs no better at balancing equity than choosing the top 500 from an unconstrained score and far worse than choosing group-specific thresholds to balance equity as shown in the main text. Notably, this result isn't a criticism of Zafar's methods when applied in appropriate contexts, but rather an indication of the limitations of current fairness-constrained methods in the resource-constrained setting we frequently encounter in machine learning to support public policy decision making.

We also explored the methods proposed by Celis\cite{ElisaCelis2019ClassificationGuarantees} and Menon and Williamson\cite{KrishnaMenon2018TheClassification}, but noticed that both methods could be shown to be equivalent to group-specific scaling or thresholding of an underlying estimate of the relevant probability distribution when applied to a single fairness metric and sub-group membership is known (notably, the method proposed by Celis\cite{ElisaCelis2019ClassificationGuarantees} seems quite flexible to balancing multiple metrics or where group membership is itself being modeled as well). For a single, monotonic metric like recall/equality of opportunity, there will be a unique balanced solution of a given total size, so any method relying on post-hoc score adjustments should yield similar results.

To see this in the context of the method described by Celis and colleagues\cite{ElisaCelis2019ClassificationGuarantees}, we can start from their observation that any equity definition balancing a confusion-matrix statistic can be represented in the following form (following the notation used in their work):

\begin{equation}
    q_{linf}^{(i)} = \frac{\alpha_{0}^{(i)} + \sum_{j\in[k]}\alpha_{j}^{(i)}P[f=1 \mid G_{i}, A_{j}^{(i)}]}
    {\beta_{0}^{(i)} + \sum_{j\in[l]}\beta_{j}^{(i)}P[f=1 \mid G_{i}, B_{j}^{(i)}]}
\end{equation}


Here, $i$ reflects subgroup membership, $A_{j}^{(i)}$ and $B_{j}^{(i)}$ are events such as $(Y=1)$ and $\alpha_{j}^{(i)}$ and $\beta_{j}^{(i)}$ are parameters defined to represent generalized form of equity function (likewise, $k,l \ge 0$ are integer values allowing the function to be written as linear combination of terms). Further, some metrics (including recall (that is, $TPR$) as we consider in this work) can be represented with a simplified linear form in which $\beta_{0}^{(i)}=1$ and $\beta_{1}^{(i)} = 0$, giving:

\begin{equation}
    q_{lin}^{(i)} = \alpha_{0}^{(i)} + \sum_{j\in[k]}\alpha_{j}^{(i)}P[f=1 \mid G_{i}, A_{j}^{(i)}]
\end{equation}

In this case, they show that fairness-improved score of an instance being classified is given by:

\begin{equation}
s_{\lambda}(x) = \eta(x) - 0.5 + 
\sum_{i \in [p]} 
        \lambda_{i} \Bigg( 
        \sum_{j\in[k]} \frac{\alpha_{j}^{(i)}}{\pi_{j}^{(i)}} \eta_{j}^{(i)}(x)
        \Bigg)
\end{equation}

Where, $\eta(x) = P[Y=1 \mid X= x]$, $\eta_{j}^{(i)}(x) = Pr[G_{i}, A_{j}^{i} \mid X=x]$
and $\pi_{j}^{(i)} = Pr[G_{i}, A_{j}^{(i)}]$, while $\lambda_{i}$ represents langrangian parameters, used to solve the optimization problem. 

For recall specifically (defined as
$ TPR = P[f=1 \mid G_{i}, Y=1] $), the sum in $q_{lin}$ has a single term ($j=1$), with $\alpha_{0}^{(i)} = 0$, $\alpha_{1}^{(i)} = 1$, $A_{1}^{(i)} = (Y=1)$. Further (and without loss of generality), for the case in which there are two sub-groups of the protected attribute (e.g., men and women), we can expand the sums and rewrite the adjusted score as:

$$
    s_{recall}(x) = P(Y=1 \mid X=x) - 0.5
    + \lambda_{0} \Bigg[\frac{1}{P[G_{0}, Y=1]} P[G_{0}, Y=1\mid X] \Bigg]
$$
\begin{equation}
    + \lambda_{1} \Bigg[\frac{1}{P[G_{1}, Y=1]} P[G_{1}, Y= 1\mid X] \Bigg]
    \qquad  \qquad  \qquad  \qquad
\end{equation}

This formulation can be quite useful for balancing recall equity in the case where group membership is itself being inferred. However, where we know $G=g$ for a given $X$, we note that this will pick out a single term as either $P[G_{1}, Y= 1\mid X] = 0$ or $P[G_{0}, Y= 1\mid X] = 0$ (note that this will be true if there were more sub-groups as well, simply adding additional terms). So, for a given example $X$, we obtain:
\begin{equation}
    s_{recall}(x) = P(Y=1 \mid X=x) - 0.5
    + \lambda_{g} \Bigg[\frac{1}{P[G=g, Y=1]} P[Y=1 \mid X] \Bigg]
\end{equation}

Rearranging leads to:
\begin{equation}
    s_{recall}(x) = \Bigg[ 1 + \frac{\lambda_{g}}{P[G=g, Y=1]}\Bigg] P(Y=1\mid X) - 0.5
\end{equation}

As it can be seen from the above equation, $1 + \lambda_{g} / P[G=g, Y=1]$ is simply a group-specific constant with no dependence on the value of $X$ beyond group membership. As such, this will result in scaling factor corresponding to rescaling of an underlying predicted score $P(Y=1 \mid X)$ around a threshold, but without reordering the score within a given subgroup itself. Further, because of the monotonically increasing nature of recall as list depth is traversed, there will be a unique (up to exact ties in the score) recall-balancing solution of a given total list size, regardless of whether that solution is obtained by setting group-specific thresholds (as in the method we make use of in the current study) or group-specific stretching scores around a fixed threshold (as we see here), so long as the within-group ranking is not reordered.

\clearpage

\begin{figure*}[htb!]
  \centering
 \includegraphics[width=\linewidth]{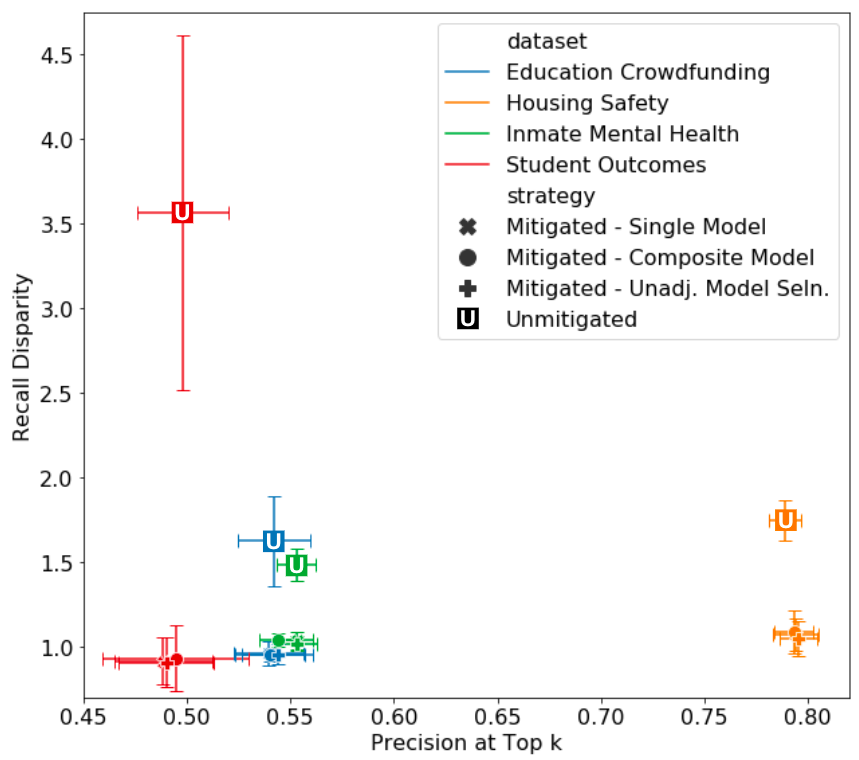}
  \caption{Comparing equity (recall disparity) and performance (precision at top $k$) metrics for different model selection strategies between different policy contexts, as in Fig. 2, including an additional model selection strategy (Mitigated - Unadj. Model Seln.) in which model specification is chosen without regard for disparities then group-specific thresholds are chosen for the selected model. Error bars reflect 95\% confidence intervals across temporal validation cohorts. This strategy performed similarly to strategy accounting for disparities in the model selection itself in all policy settings considered.}
  \label{fig:compare-with-unadj}
\end{figure*}

\clearpage

\begin{figure*}[htb!]
  \centering
  \includegraphics[width=0.8\linewidth]{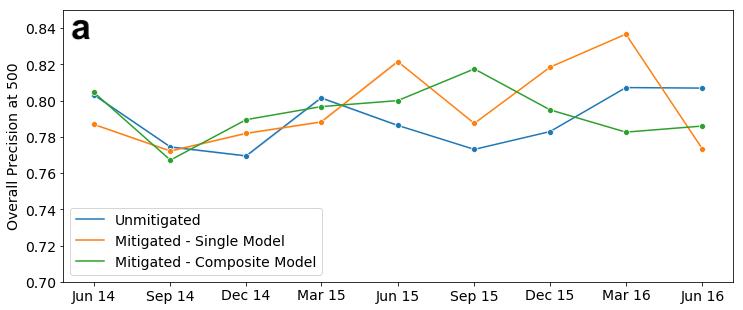}
  \includegraphics[width=0.8\linewidth]{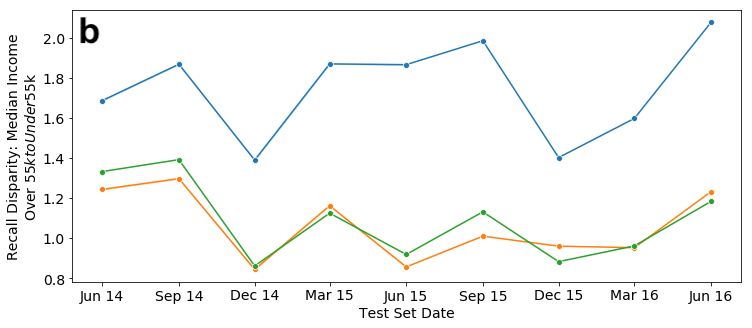}
  \caption{Comparing model precision (a) and disparity (b) metrics over time for different model selection strategies in the Housing Safety policy context.}
  \label{fig:san-jose}
\end{figure*}

\clearpage

\begin{figure*}[htb!]
  \centering
  \includegraphics[width=0.8\linewidth]{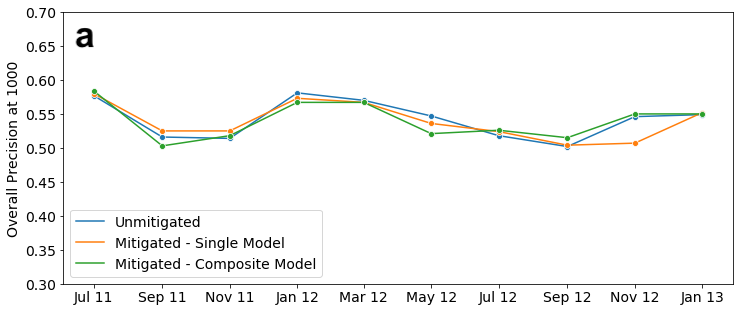}
  \includegraphics[width=0.8\linewidth]{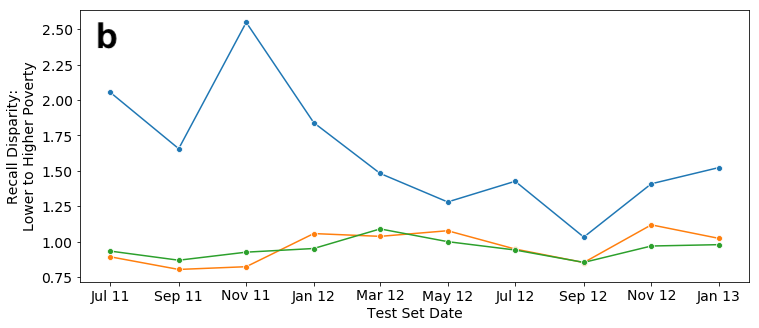}
  \caption{Comparing model precision (a) and disparity (b) metrics over time for different model selection strategies in the Education Crowdfunding policy context.}
  \label{fig:donors-choose}
\end{figure*}

\clearpage

\begin{figure*}[htb!]
  \centering
  \includegraphics[width=0.8\linewidth]{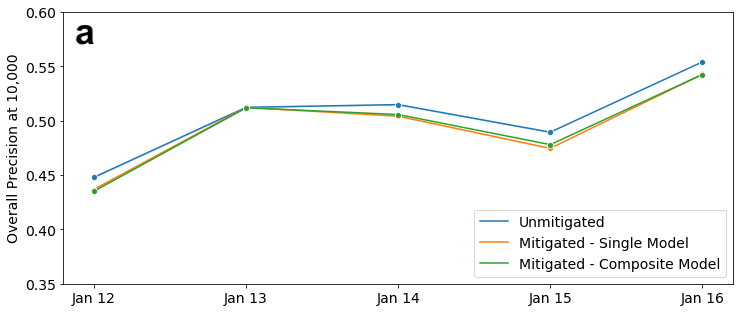}
  \includegraphics[width=0.8\linewidth]{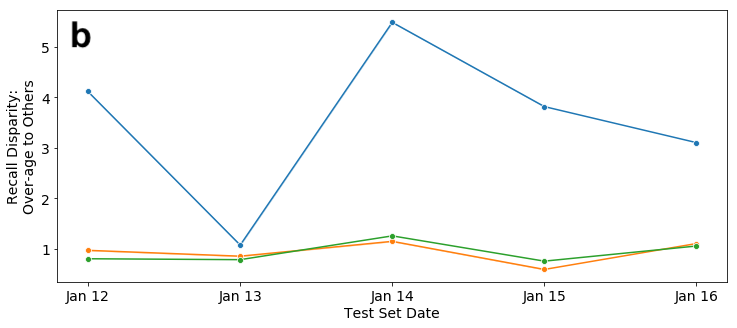}
  \caption{Comparing model precision (a) and disparity (b) metrics over time for different model selection strategies in the Student Outcomes policy context.}
  \label{fig:el-salvador}
\end{figure*}

\clearpage

\begin{figure*}[htb!]
  \centering
  \includegraphics[width=0.8\linewidth]{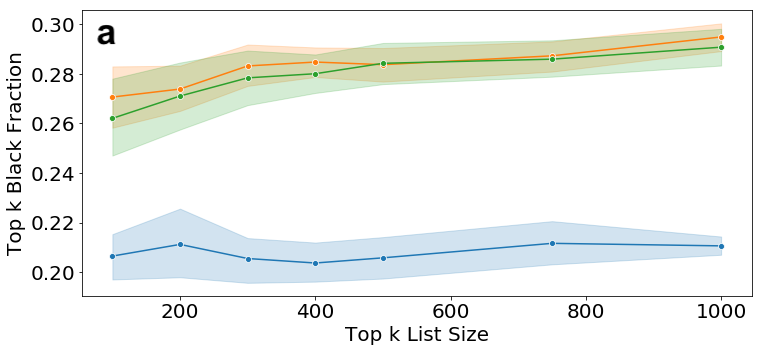}
  \includegraphics[width=0.8\linewidth]{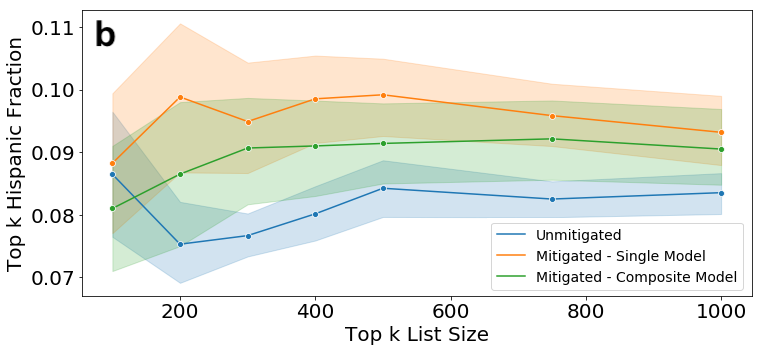}
  \caption{Fraction of the selected ``top-k'' list that is African American (a) or Hispanic (b) by list size in the Inmate Mental Health setting (see Fig. 4a-c). Shaded intervals reflect variation across temporal validation splits.}
  \label{fig:list-size}
\end{figure*}

\clearpage

\begin{figure*}[htb!]
  \centering
 \includegraphics[width=0.8\linewidth]{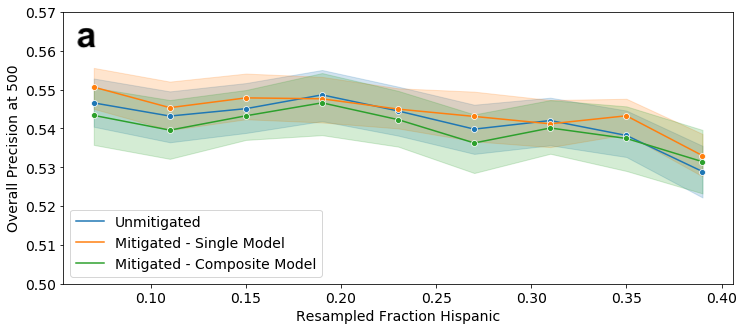}
 \includegraphics[width=0.8\linewidth]{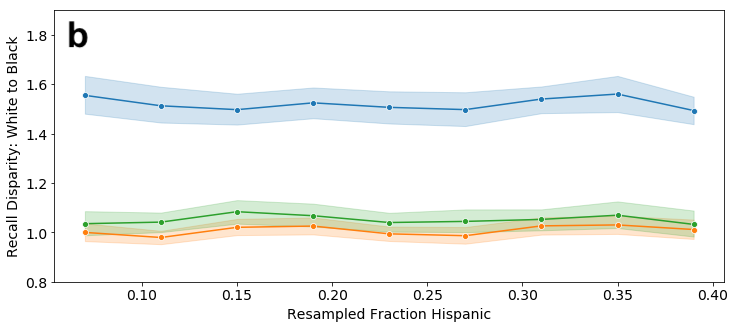}
  \caption{Precision (a) and white-to-Black disparities (b) by re-sampled Hispanic fraction from the experiment shown in Fig. 4d. Shaded intervals reflect variation across bootstrap samples and temporal validation splits.}
  \label{fig:bootstrap}
\end{figure*}

\clearpage

\begin{figure*}[htb!]
  \centering
  \includegraphics[width=0.5\linewidth]{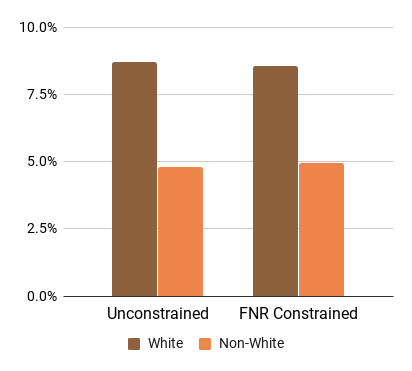}
  \caption{Comparison of recall disparity after selecting the top 500 individuals from unconstrained and False Negative Rate (FNR)-constrained models following the methods described in Zafar et al. Results from the Inmate Mental Health policy context.}
  \label{fig:regularized}
\end{figure*}



\clearpage

\begin{table}[htb!]
 \caption{Model Hyperparameter Grid for Inmate Menthal Health Policy Setting}
  \centering
  \begin{tabularx}{\linewidth}{ l l X }
    \hline
    Estimator     & Hyperparamter & Grid Search Values  \\
    \hline
    \verb|RandomForestClassifier| & max\_features & sqrt \\
    & criterion & gini, entropy \\
    & n\_estimators & 100, 1000, 5000 \\
    & min\_samples\_split & 10, 25, 100 \\
    & max\_depth & 5, 10, 50 \\
    \hline
    \verb|ScaledLogisticRegression| & penalty & l1, l2 \\
    & C & 0.001, 0.1, 1, 10 \\
    \hline
    \verb|PercentileRankOneFeature| & feature & Jail bookings last 1 or 5 years \\
    \hline
    \verb|SimpleThresholder| & rules & Any mental health history, \\
    & & Jail releases in last: 1, 3, 6 months, or 1 year \\
    & logical\_operator & and \\
    \hline
  \end{tabularx}
  \label{tab:joco-model-grid}
\end{table}

\clearpage

\begin{table}[htb!]
 \caption{Model Hyperparameter Grid for Education Crowdfunding Policy Setting}
  \centering
  \begin{tabularx}{\linewidth}{ l l X }
    \hline
    Estimator     & Hyperparamter & Grid Search Values  \\
    \hline
    \verb|DecisionTreeClassifier| & criterion & gini \\
    & min\_samples\_split & 2, 5, 10, 100, 1000 \\
    & max\_depth & 1, 5, 10, 20, 50, 100 \\
    \hline
    \verb|RandomForestClassifier| & max\_features & sqrt \\
    & criterion & gini, entropy \\
    & n\_estimators & 100, 1000, 2000, 3000 \\
    & min\_samples\_split & 10, 50\\
    & max\_depth & 10, 50, 100 \\
    \hline
    \verb|ExtraTreesClassifier| & max\_features & log2 \\
    & criterion & entropy \\
    & n\_estimators & 10, 50, 1000 \\
    & min\_samples\_split & 5, 25, 50 \\
    & max\_depth & 20, 50, 100 \\
    \hline
    \verb|ScaledLogisticRegression| & penalty & l1, l2 \\
    & C & 0.0001, 0.001, 0.01, 0.1, 1, 10 \\
    \hline
    \verb|AdaBoostClassifier| & n\_estimators & 500,1000 \\
    \hline
  \end{tabularx}
  \label{tab:dc-model-grid}
\end{table}

\clearpage

\begin{table}[htb!]
 \caption{Model Hyperparameter Grid for Housing Safety Policy Setting}
  \centering
  \begin{tabularx}{\linewidth}{ l l X }
    \hline
    Estimator     & Hyperparamter & Grid Search Values  \\
    \hline
    \verb|DecisionTreeClassifier| & criterion & gini, entropy \\
    & min\_samples\_split & 10, 20, 50, 100 \\
    & max\_depth & 1, 2, 3, 5, 10, 20, 50 \\
    \hline
    \verb|RandomForestClassifier| & max\_features & sqrt, log2 \\
    & criterion & gini, entropy \\
    & n\_estimators & 100, 1000, 5000 \\
    & min\_samples\_split & 10, 20, 50, 100 \\
    & max\_depth & 2, 5, 10, 20, 50, 100 \\
    \hline
    \verb|ExtraTreesClassifier| & max\_features & sqrt, log2 \\
    & criterion & gini, entropy \\
    & n\_estimators & 100, 1000, 5000 \\
    & min\_samples\_split & 10, 20, 50, 100 \\
    & max\_depth & 2, 5, 10, 50, 100 \\
    \hline
    \verb|ScaledLogisticRegression| & penalty & l1, l2 \\
    & C & 0.001, 0.01, 0.1, 1, 10 \\
    \hline
    \verb|PercentileRankOneFeature| & feature & Days since last routine inspection \\
    \hline
  \end{tabularx}
  \label{tab:sj-model-grid}
\end{table}

\clearpage

\begin{table}[htb!]
 \caption{Model Hyperparameter Grid for Student Outcomes Policy Setting}
  \centering
  \begin{tabularx}{\linewidth}{ l l X }
    \hline
    Estimator     & Hyperparamter & Grid Search Values  \\
    \hline
    \verb|DecisionTreeClassifier| & criterion & gini \\
    & min\_samples\_split & 2, 5, 10, 100, 1000 \\
    & max\_depth & 1, 5, 10, 20, 50, 100 \\
    \hline
    \verb|RandomForestClassifier| & max\_features & sqrt \\
    & criterion & gini \\
    & n\_estimators & 100, 500 \\
    & min\_samples\_split & 2, 10\\
    & class\_weight & \textasciitilde, balanced subsample, balanced \\
    & max\_depth & 5, 50, 100 \\
    \hline
    \verb|ExtraTreesClassifier| & max\_features & log2 \\
    & criterion & gini  \\
    & n\_estimators & 100 \\
    & max\_depth & 5, 50 \\
    \hline
    \verb|ScaledLogisticRegression| & penalty & l1, l2 \\
    & C & 0.0001, 0.001, 0.01, 0.1, 1, 10 \\
    \hline
  \end{tabularx}
  \label{tab:mined-model-grid}
\end{table}

\clearpage

\renewcommand\refname{Supplementary References}